# Wide and Deep Learning for Peer-to-Peer Lending

Kaveh Bastani[1]*, Elham Asgari[2], Hamed Namavari[3]

[1]Unifund CCR, LLC, Cincinnati, OH
[2]Pamplin College of Business, Virginia Polytechnic Institute, Blacksburg, VA
[3]Economics, College of Business, University of Cincinnati, Cincinnati, OH

**Abstract -** This paper proposes a two-stage scoring approach to help lenders decide their fund allocations in the peer-to-peer (P2P) lending market. The existing scoring approaches focus on only either probability of default (PD) prediction, known as credit scoring, or profitability prediction, known as profit scoring, to identify the best loans for investment. Credit scoring fails to deliver the main need of lenders on how much profit they may obtain through their investment. On the other hand, profit scoring can satisfy that need by predicting the investment profitability. However, profit scoring completely ignores the class imbalance problem where most of the past loans are non-default. Consequently, ignorance of the class imbalance problem significantly affects the accuracy of profitability prediction. Our proposed two-stage scoring approach is an integration of credit scoring and profit scoring to address the above challenges. More specifically, stage 1 is designed as credit scoring to identify non-default loans while the imbalanced nature of loan status is considered in PD prediction. The loans identified as non-default are then moved to stage 2 for prediction of profitability, measured by internal rate of return. Wide and deep learning is used to build the predictive models in both stages to achieve both memorization and generalization. Extensive numerical studies are conducted based on real-world data to verify the effectiveness of the proposed approach. The numerical studies indicate our two-stage scoring approach outperforms the existing credit scoring and profit scoring approaches.

**Keywords –**Wide and Deep Learning, Peer-to-peer Lending, Credit scoring, Profit scoring

## 1. Introduction

The P2P lending market consists of individuals who lend to and borrow from each other using an Internet-based platform. This platform receives loan requests from borrowers and provides lenders with investment opportunities to fund these requests. Lenders review borrowers' applications and eventually may approve to fund the loans partially. As lenders only receive their principal back if borrowers pay loans in full, lending decision involves financial risk. More specifically these loans are unsecured, and lenders should bear the full risk of losing money if borrowers default on the loans.

---

[1] Corresponding author, kaveh@vt.edu, 10625 Techwoods Circle, Cincinnati, OH 45242

To help lenders manage the above risk, it would be crucial to determine the level of risk associated to each loan. The risk level is typically defined in terms of how likely the borrower defaults on the loan, also known as probability of default (PD). The riskier loans encompass larger PD while safer loans contain lower PD. The loan's PD is not known at the time of investment, but it can be predicted using multiple sources of information available on the loan and the borrower. The factors explaining PD have been reviewed in recent studies (Serrano-Cinca et al. 2015). These factors include variables such as loan purpose, FICO score, borrower's assets, employment status, etc. Using the above factors, there have been several analytical approaches proposed to predict PD in P2P lending (Serrano-Cinca et al. 2015; Malekipirbazari et al., 2015; Guo et al., 2016). These studies focus on developing a classifier using machine learning algorithms to predict borrower's PD. Given the predicted PD, a loan can be classified into either (1) non-default if PD is lower than a predefined threshold (e.g. 0.5), or (2) default otherwise. These approaches which are also known as credit scoring are used to score the loans based on their predicted PD. Lower scores are given to the loans with higher PD and vice versa. Subsequently, the lenders might be able to reduce the risk of investment by funding the loans with higher scores.

Although credit scoring approaches have shown interesting results in lowering the risk of investment, they may not fully address the true objective of lenders in P2P lending. The lenders not only care about the loan's PD, but also the profit that they generate from their investment. Loans with higher risks typically include higher interest rates, hence, the investor will be able to gain more profit by funding those loans if the borrowers successfully pay off the loans. For example, a business loan might be determined more risky (higher PD) than an auto loan (lower PD) but funding a business loan could potentially result into higher profit than that of an auto loan. Serrano-Cinca and Gutiérrez-Nieto (2016) proposed a profit scoring approach to predict the profitability of the loans. Internal rate of Return (IRR) was used as a measure of profitability in their work. IRR is a common financial formula that can be easily used to compute the effective interest rate of the lenders. The loans are assigned to the borrowers with corresponding interest rates. However, the effective interests that lenders receive might be different from the interest rates that borrowers pay. For example, a borrower is assigned a loan with an annual interest rate of 5% with 3-year term. If the borrower pays off the loan in 3 years, the lender's effective interest rate would be equivalent to 5%. However, if the borrower pays off the loan earlier than 3 years, the IRR would be lower than 5%. The IRR could even receive negative value. If the borrower defaults on the loan, depending on the amount of loan delinquency, the IRR could theoretically become a value ranging from negative infinity to a small negative value. The proposed profit scoring approach in (Serrano-Cinca and Gutiérrez-Nieto, 2016) can be used to score the loans based on their predicted IRRs. Subsequently, the lenders might select the loans with the highest IRR.

As discussed above, profit scoring and credit scoring conceptually work in opposite favor of each other. In other words, profit scoring focuses on the loans that are highly profitable and ignores the PD associated with those loans, on the other hand, credit scoring focuses on the loans with the lowest PD and disregards the profitability of the loans. In this paper, we propose a scoring approach based on an integration of credit scoring and profit scoring. Specifically, a two-stage scoring approach is developed; stage 1 predicts PD to determine non-default loans from the listings[2] and stage 2 is used to predict the profitability of the loans identified as non-default in stage 1. The proposed scoring approach is capable of taking into account both PD and profitability in scoring the loans. Subsequently, the lenders might be able to select the loans with the highest profitability whose PD have been thoroughly analyzed.

Our two stage approach is based on wide and deep learning algorithm developed by Google scientists (Cheng et al., 2016). Wide and deep learning is capable of achieving both memorization and generalization. Memorization is defined as learning the frequent interactions of features from the historical data. Generalization, on the other hand, refers to exploiting new feature interactions that have never or rarely occurred in historical data.[3] As opposed to other machine learning algorithms, such as regression, deep learning and random forest, that may only take into account either generalization or memorization, wide and deep learning focuses on both which makes it a strong candidate for developing our proposed two-stage scoring approach (more details on competence of wide and deep learning are discussed in Section 3). Stage 1 of our proposed approach is formulated as a classification problem aiming to classify the loans into either non-default or default. Stage 2 is formulated as a regression problem aiming to predict the IRR of the loans. Wide and deep learning is applied to solve both classification and regression problems formulated in stage 1 and stage 2, respectively. To validate the effectiveness of the proposed approach, extensive experiments are conducted using real-world data from Lending Club which is one of the largest P2P lending marketplaces in the US. The empirical results indicate that the proposed scoring approach effectively outperforms the existing credit scoring and profit scoring approaches.

---

[2]Listings include the loan requests from all the borrowers. The lenders are allowed to review the listings and decide which loans they are interested to invest in and how much they fund those loans.

The remainder of the paper is organized as follows: Section 2 provides a review on the existing credit scoring and profit scoring approaches in the context of P2P lending. Section 3 is devoted to introduce wide and deep learning. Section 4 proposes our two-stage scoring approach using wide and deep learning. Section 5 demonstrates the effectiveness of the proposed approach using experimental studies. Finally, Section 6 concludes the paper.

**2. Literature review**

Credit scoring is formulated as a classification problem with a binary dependent variable which assigns zero to default loans and one to non-default loans. The aim of credit scoring is to classify a loan into either default or non-default by predicting PD. There are various sources of information available on loan and borrowers which can be used to predict PD in P2P lending. Serrano-Cinca et al. (2015) reviewed the determinant factors of PD and identified their importance in PD prediction.[4] These factors were grouped into 5 categories including (1) loan characteristics (e.g. loan amount, purpose), (2) borrower characteristics (e.g.

---

[3] The terms features and factors are used interchangeably through the paper to refer to the variables used for prediction of the outcome (e.g. PD or IRR).
[4] All these variables are available online and the prospective lenders would be able to utilize these variables for making their investment decisions.

annual income, housing status), (3) borrower assessment (e.g. FICO score, interest rate, grade), (4) borrower indebtedness (e.g. debt to income), and (5) credit history (e.g. number of delinquencies, revolving utilization). Given the determinant factors, they proposed a logistic regression model to predict PD. They showed that factors such as grade and FICO are the main variables in explaining PD, however, the prediction performance of the model can be improved by adding other determinant factors.

Malekipirbazari and Aksakalli (2015) proposed a random forest (RF) based classification approach to identify the loan status of the borrowers. They compared the performance of RF with other machine learning algorithms namely, support vector machine (SVM), k-nearest neighborhood (k-NN), and logistic regression, and showed that RF outperforms other classifiers in PD prediction. Guo et al. (2016) proposed an instance-based model to predict the return rate and risk of loans in P2P lending. The return rate of a new loan was predicted as a weighted average of the return rate of historical instances (i.e. past loans), where the weights were estimated based on similarities of the loans. The similarities of the loans were defined in terms of Euclidean distance of the loans' PD. More specifically PD of each loan was first determined using logistic regression, and then the loans' PD Euclidean distances were computed.[5] Subsequently to obtain the optimal weights, kernel regression was used to capture the nonlinear relationship between the computed raw PD distances and the weights. Jiang et al. (2018) proposed a topic modeling based model to predict PD by analyzing descriptive text concerning loans. Topic modeling has been shown as an effective techniques to analyze textual data in various applications (Asgari et al. 2017). Their empirical results using real-word data from a major P2P lending platform in China show the effectiveness of their proposed method in PD prediction by analyzing both text and non-text based features. Kim and Cho (2018) developed a deep dense convolutional networks for default prediction in P2P social lending. Their numerical results validate that the proposed method automatically extracts useful features from Lending Club data and is effective in PD prediction. The credit scoring approaches reviewed above have been reported in the context of P2P lending.

---

[5]If a new loan has less PD distance with a past loan, it is assumed that these two loans are less similar, and intuitively corresponding weight should be low.

The interested readers are suggested to refer to (Lessmann et al., 2015) for a detailed review of credit scoring approaches used in other areas of consumer credit such as credit card and mortgages (So et al., 2014).

Over the past few years, the focus of credit lenders have changed from minimizing the risk of defaulting consumers (i.e. credit scoring) to maximizing the profit margins. As a result, profit scoring approaches have received a large amount of attention recently. Various profit scoring approaches have been proposed in the literature of consumer credit risk. These approaches have been based on analytical techniques including Markov chain modeling (Thomas et al., 2002a; Thomas et al., 2002b), survival analysis (Narain 1992; Banasik et al., 1999; Sanchez-Barrios et al. 2016), expected profit maximization (Finlay 2008; Finlay 2020, Stewart 2011; Verbraken 2014), and regression (Buckley and James 1979; Lai and Ying, 1994). Markov chain techniques have been used to build stochastic models of complex situations and consumer behaviors. Survival analysis techniques such as proportional hazards (Breslow 1975) and accelerated life models (Bradburn 2003) have been used to estimate the long term profitability behavior of the consumer. Expected profit maximization approaches have been developed based on novel profitability performance measures of consumers. Finlay (2010) proposed a profitability score of expected return and loss, and applied genetic algorithm (GA) to optimize the profit gains. Verbraken et al (2014) developed a profit-based classification performance measure based on expected maximum profit (EMP). The proposed measure was further used to find the optimal cut-off threshold for implementation of classification algorithms. The regression approach formulates profit as a dependent variable which is set to be predicted from a group of determinant factors.

Although numerous profit scoring approaches have been reported in the literature of consumer credit risk, a very few of these works have been focused on the context of P2P lending. One of those instances is Serrano-Cinca and Gutiérrez-Nieto (2016) wherein the authors proposed IRR as a measure of profitability of loans, and studied determinant factors of IRR. These factors were very similar to the determinant factors of PD studied by Serrano-Cinca et al. (2015) (more details on these variables are provided in Section 4). They further developed a decision tree model to predict IRR using determinant factors. Decision trees are capable of capturing non-linear relations between determinant factors and IRR, while producing a set of decision rules that are very easy to assimilate. They conducted numerical studies on the Lending Club dataset to verify the effectiveness of their proposed approach. The authors compared the performance of the proposed profit scoring approach with a credit scoring approach built on logistic regression. Their numerical studies indicated that the lenders will be able to obtain higher IRR using profit scoring rather than credit scoring.

In spite of profitability improvements achieved by Serrano-Cinca and Gutiérrez-Nieto (2016), their proposed approach fails to consider the fact that P2P lending suffers from imbalanced distribution of loans. From historical data in the P2P lending market, almost 15% of the past loans are defaulted and 85% are non-defaulted.[6] Obviously this results into an imbalanced dataset, where the distribution of the loans is not balanced (please refer to Figure 1). A predictive model that is built on an imbalanced dataset will be biased towards the majority classes/values (i.e. non-default class or positive IRR), and fails to accurately predict the minority classes/values (default class or negative IRR). The class imbalance problem in P2P lending is primarily caused by unequal distribution of loan status which can be addressed in PD prediction. The work of Ref. Serrano-Cinca and Gutiérrez-Nieto (2016) completely ignores PD associated to the loans, and, instead, directly predicts IRR. Hence, their approach does not consider the imbalanced nature of the loans.

To address the shortcomings of the reviewed credit scoring and profit scoring approaches, a two-stage scoring approach is proposed in this paper. The first stage is designed to identify non-default loans while the imbalanced nature of loans is taken into account in the PD prediction. The loans identified as non-default are then moved to stage 2 for the IRR prediction. Wide and deep learning is used to build the predictive models in the proposed approach, to achieve both memorization and generalization. Section 3 introduces wide and deep learning, and subsequently the proposed scoring approach using wide and deep learning is presented in Section 4.

---

[6]The default rate mentioned here refers to the Lending Club loan and borrowers data. But it would be very similar in other P2P platforms such as Prosper as well.

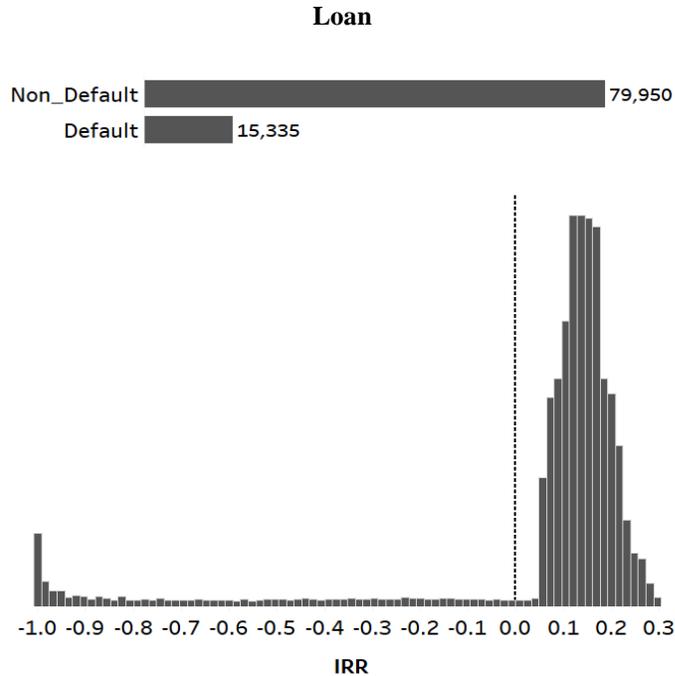

Figure 1. Loan status and IRR Histograms are presented. The issue of imbalance dataset exists whether loan status or IRR is taken into account as an output of scoring approaches. In case of loan status, almost 85% (79,950 in the Lending Club dataset) of the loans are non-default and 15% (15,335 in the Lending Club dataset) are default. Similarly in case of IRR, almost 85% of the loans have positive IRR and 15% have negative IRR.

## 3. Wide and deep learning

Google scientists developed and commercialized wide and deep learning algorithm for mobile application recommender systems on the Google Play store (Cheng et al., 2016). Wide and deep learning allows them to recommend their users a diverse variety of mobile applications (generalization), yet customized to user information (memorization). The main contribution of wide and deep learning is, in fact, in integration of linear model and neural networks to achieve both memorization and generalization, respectively. It would be beneficial to introduce the concepts of memorization and generalization using a practical example. Figure 2 shows histograms of two categorical features, namely, FICO score and the loan purpose (these two features are among the determinant factors of PD and IRR). A partition of interactions of these two features is also presented in this figure.

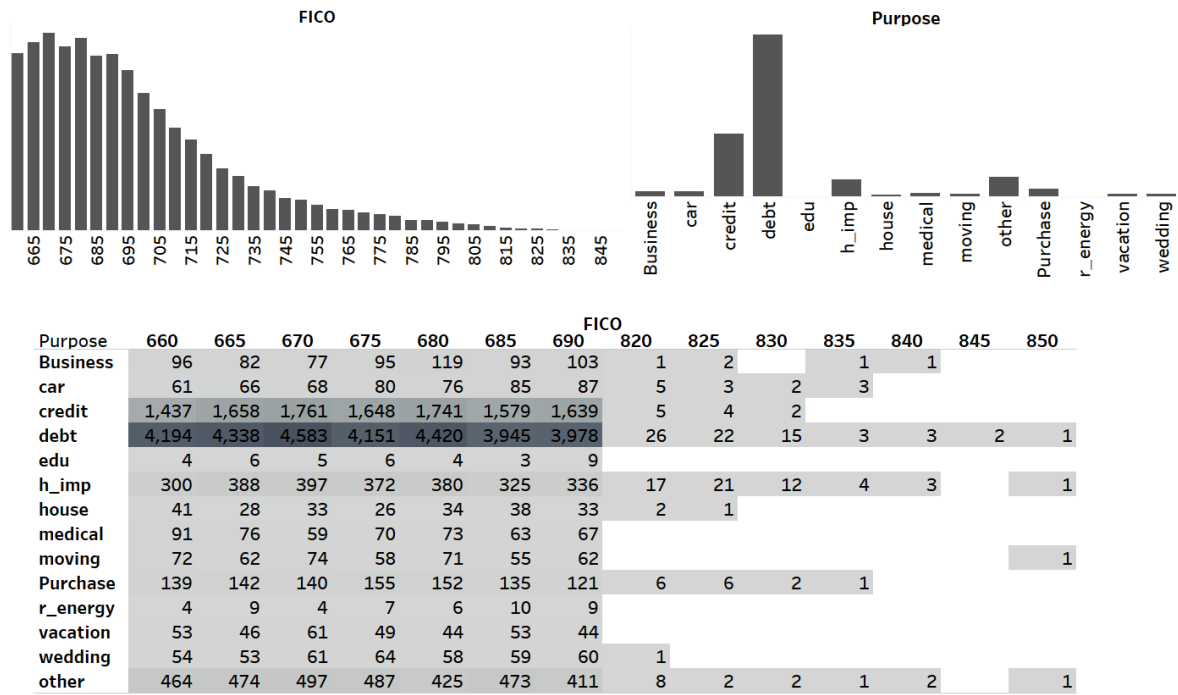

Figure 2. Top section shows histogram of the FICO score, and the loan purpose distribution. Bottom section presents the interactions (co-occurrences) of FICO score and the loan purpose (only a few of FICO scores are presented for the sake of space). From the bottom chart it is obvious that high FICO scores with loan purpose rarely co-occurred in the past loans.

Let $\mathbf{x}_1$ represent FICO score which is a measure of consumer credit risk based on credit report. FICO score is a categorical variable with 35 levels, ranging from 660-850 (with increment step of 5)[7]. Let $\mathbf{x}_2$ be the loan purpose which indicates the reason that the consumer borrows the loan. The loan purpose includes 14 levels such as auto loan, debt consolidation, business, etc. In general, a categorical feature is represented as a one-hot vector where all elements are 0 except only one element that relates to the state of the feature, which is given a value of 1. For example, if FICO score is "660", then $\mathbf{x}_1 = [1,0,0,0,\ldots,0,0]^T$, where the element related to "660" is 1 and other elements are 0. Similarly, if the loan purpose is "auto-loan", then $\mathbf{x}_2 = [1,0,0,0,\ldots,0,0]^T$, where the element related to "auto-loan" is 1 and other elements are 0.

**3.1.1. Memorization through wide learning**

Memorization emphasizes on frequent co-occurrences of the features in the past, and exploits their interactions in training the model. For example FICO score of "660" has co-occurred frequently with the loan purpose of "debt consolidation". The interaction term of FICO and the loan purpose enables us to account for the frequent co-occurrences of these two features. Indeed, memorization can be effectively captured by adding the interaction term to a wide learning model. Wide learning is a generalized linear model (such as logistic regression or linear regression) in the form of

$$y = \mathbf{w}^T \mathbf{x} + b \quad (1)$$

---

[7] Generally FICO score is an ordinal variable in range if 300-850. However, FICO scores in the Lending Club datasets are represented as a factor of 5, and located in the range of 660-850. Therefore, FICO is considered as a categorical variable. More discussion on FICO variable is provided in Section 5.1.2 and Appendix II.

, where y is the prediction output (PD, and IRR in our case), $\mathbf{x} = [\mathbf{x}_1, \mathbf{x}_2, \ldots, \mathbf{x}_m]$ is a vector of features, $\mathbf{w} = [\mathbf{w}_1, \mathbf{w}_2, \ldots, \mathbf{w}_m]$ denotes model parameters, and b is the bias (please refer to Figure 3). The feature vector $\mathbf{x}$ includes the raw features (e.g. FICO ($\mathbf{x}_1$) and the loan purpose($\mathbf{x}_2$)), and the transformed features (e.g. interaction of FICO and loan purpose).

Similar to categorical features, the interaction term is also defined as a one-hot vector where all elements are 0, except the element associated to the status of the constituent features. For example, if the constituent features are FICO score of "660" and the loan purpose of "debt consolidation", the interaction term denoted by AND ($\mathbf{x}_1$, $\mathbf{x}_2$) is represented as a one-hot vector $[0,0,\ldots,0,1,0,\ldots,0]$ with dimension of $|\mathbf{x}_1|.|\mathbf{x}_2|$ ($|\cdot|$ is the cardinality of a vector), where the element related to (FICO score of "660" and purpose of "debt consolidation") is 1 and all others are 0. The interaction term captures correlations of the constituent features and adds nonlinearity to the wide learning model.

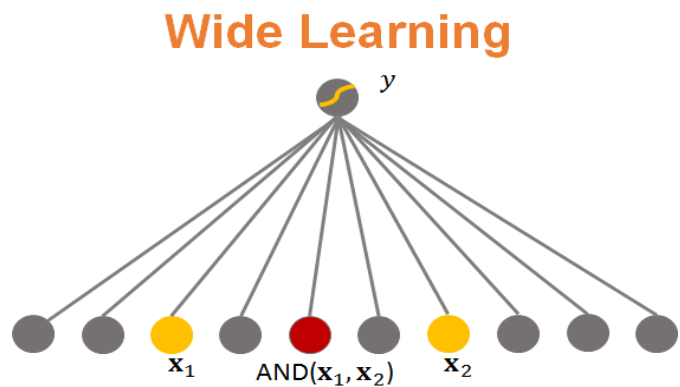

Figure 3. Illustration of the wide model. The raw features (e.g. FICO $\mathbf{x}_1$ and the loan purpose $\mathbf{x}_2$) along with the transformed features (e.g. interaction term AND ($\mathbf{x}_1, \mathbf{x}_2$)) are the inputs, and y is the prediction output (e.g. the loan status, or IRR). Note that in case of the loan status, the output y is transformed using $\frac{1}{1+e^{-y}}$ to create values between 0 and 1 in order to account for PD.

**3.1.2 Generalization through deep learning**

Generalization refers to exploiting new feature interactions that have never or rarely occurred in historical data. In fact, in case of low or no co-occurrences of the features, the interaction term does not provide any useful information as no data exists for training the model. For example, FICO score of "840" has rarely co-occurred with the loan purpose of "business", or FICO score of "850" has never co-occurred with the loan purpose of "auto-loan" (please refer to Figure 2). Therefore, wide learning cannot generalize prediction to the scenarios that have never or rarely been occurred in historical data, and consequently may not be able to accurately predict (e.g. loan status or IRR) given these scenarios. On the other hand, deep neural networks can generalize to unseen feature interactions. Deep learning allows learning a low dimensional dense embedding vector for each categorical feature with less efforts of feature engineering. Embedding vector is a real-valued vector that is obtained by conversion of a categorical feature from its high dimensional sparse space (i.e. one-hot vector representation) into a low dimensional continuous space. In the embedding space, we can generalize our prediction given any feature interactions, especially the missing interactions in the data (e.g. "850" FICO score and the auto-loan borrowing purpose), which could have not been possible by wide learning.

Deep learning model is presented in Figure 4. The model consists of embedding vectors along with neural network hidden layers. Each hidden layer is defined as below:

$$\boldsymbol{a}^{l+1} = f(\mathbf{W}^l \boldsymbol{a}^l + \mathbf{b}^l) \qquad (2)$$

, where $l$ denotes the layer number, $f(\cdot)$ is an activation function (ReLU in most cases), $\boldsymbol{a}$ is the activation, $\mathbf{W}$ is the model weights, and $\mathbf{b}$ is the bias at layer $l$. The unknown variables include $\mathbf{W}^l$, $\mathbf{b}^l$, and the embedding vectors; these variables are all randomly initialized (e.g. random draws from Gaussian distribution), and will be learned to minimize the loss function during the training procedures.

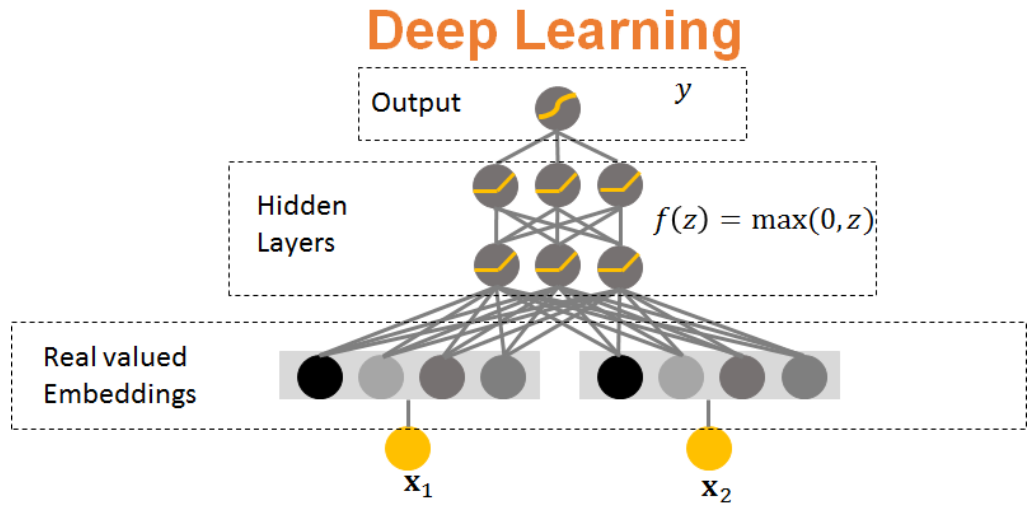

Figure 4. Illustration of Deep model. The raw features (e.g. FICO $\mathbf{x}_1$ and the loan purpose $\mathbf{x}_2$) along with their real-valued embeddings, and hidden layers $a^{l+1}$ are presented in this figure. The notation $f(\cdot)$ represents activation function (e.g. ReLU), and y is the prediction output (e.g. the loan status, or IRR). Note that in case of the loan status, the output y is transformed using $\frac{1}{1+e^{-z}}$ to create values between 0 and 1 in order to account for PD.

Training in deep learning is carried out by backpropagating the gradients from the loss function of the output (y) to all the hidden layers, and embedding vectors using mini-batch stochastic gradient descent (Bottou 2010). Various loss functions can be used in the training step. For example, cross entropy is common for classification problems, and mean squared errors is common for regression problems.

**3.1.3 Generalization and memorization through wide and deep learning**

Integration of wide leaning and deep learning results into a unified model that is capable of achieving both memorization and generalization. The wide and deep learning accounts for diverse yet relevant interaction scenarios and avoid over-generalization which is typically obtained by deep learning. The structure of wide and deep learning is presented in Figure 5. The wide and deep components are combined using a weighted sum of their outputs along with an activation function (e.g. logit function for classification and linear function for regression). Joint training of wide and deep components is carried out by backpropagating the gradients from the model output to both wide and deep components using mini-batch stochastic optimization.

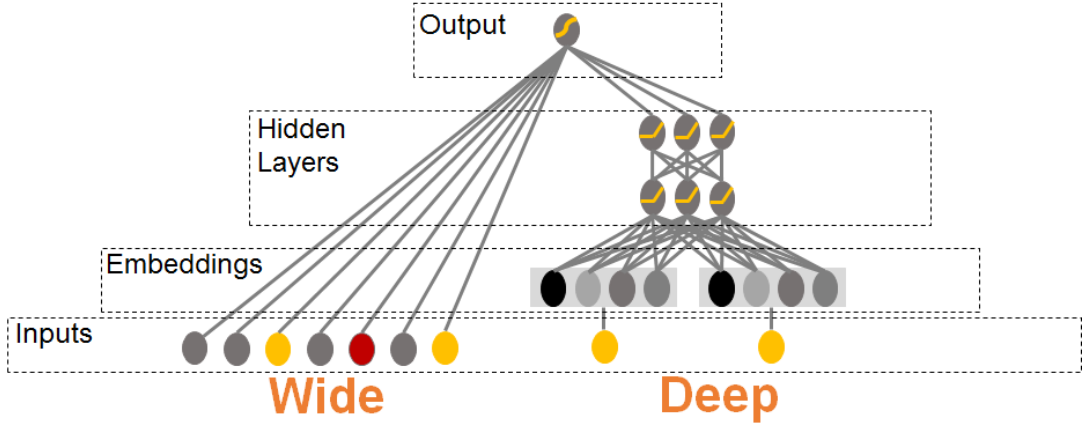

Figure 5. Illustration of the wide and deep model, which is an integration of wide component and deep component to achieve both memorization and generalization.

Once the model is trained, it predicts the output by using the following equations; for classification (classify loan status by the PD prediction in our case)

$$p(y = 1|\mathbf{x}) = \sigma(\mathbf{W}_{wide}\mathbf{x} + \mathbf{W}_{deep}\mathbf{a}^{l_f} + b) \quad (3)$$

, and for regression (the IRR prediction in our case)

$$y = \mathbf{W}_{wide}\mathbf{x} + \mathbf{W}_{deep}\mathbf{a}^{l_f} + b \quad (4)$$

, where $y$ denotes the model prediction (e.g. $y = 1$ shows the default loan in the PD prediction in stage 1, and y shows the IRR in stage 2), $\sigma(\cdot)$ is the logit function (i.e. $\frac{1}{1+e^{-z}}$), $\mathbf{W}_{wide}$ denotes all the weights for wide component, and $\mathbf{W}_{deep}$ denotes the weights applied to the last activations $\mathbf{a}^{l_f}$, and $b$ is the bias. In this paper, wide and deep learning is used to build the predictive models in the proposed two-stage scoring approach. The predictive model in stage 1 uses Eq. (3) for PD prediction. The predictive model in stage 2 uses Eq. (4) for the IRR prediction. The details of the proposed approach are discussed in the following section.

**4. Two-stage scoring approach using wide and deep learning**

The overview of the proposed two-stage scoring approach is presented in Figure 6. Stage 1 predicts PD of the loans using PD features studied in Serrano-Cinca et al. (2015).There is a checkpoint between stage 1 and stage 2 in order to evaluate the loan status given the predicted PD. Loans whose PD are larger than a threshold parameter $\gamma$ are identified as default loans, and filtered out from further analysis. The remaining loans with predicted non-default status are moved to stage 2. In stage 2, IRR of non-defaults loans are predicted using IRR features studied in Serrano-Cinca, and Gutiérrez-Nieto (2016). Given the predicted IRR, a lender would be able to select the loans with the highest IRR and invest on those loans. The details of the predictive models in stage 1 and stage 2 are presented as follows.

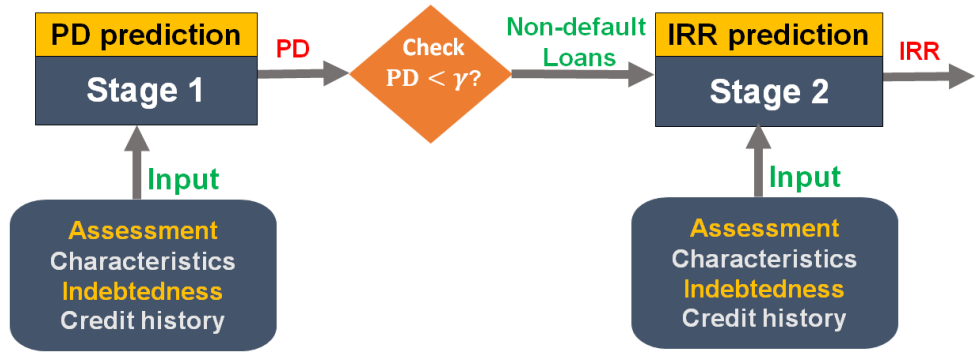

Figure 6. Overview of the proposed two-stage scoring approach

**4.1 Stage 1–PD prediction**

**4.1.1. PD determinant features**

In practice, a predictive model (classifier) is built on a group of features that are important in predicting the model output (e.g. PD). In this paper, the predictor features of PD are adopted from the Lending Club data[8] as presented in Table 1. These features are similar to the PD determinants studied in Serrano-Cinca et al. (2015). Table 1 presents these features which are grouped into 5 categories: (1) loan characteristics, (2) borrower characteristics, (3) borrower assessment, (4) borrower indebtedness, and (5) credit history. The above features form the independent variables, while the dependent variable is the status of the loan (i.e. default or non-default loan). The dependent variable suffers from the problem of class imbalance which is discussed in the following. For detailed descriptive statistics on these features please refer to Appendix I.

---

[8]To have a fair comparison with other scoring approaches such as Serrano-Cinca and Gutiérrez-Nieto (2016), the predictive models in this paper are also built on the Lending Club data.

Table 1. The features used in both PD predictive and IRR predictive modeling

| Grade | Lending Club assigns each loan a grade from A to G using a proprietary algorithm based on the loan and borrower information. Grade A is the safest loan and Grade G is the riskiest loan. |
|---|---|
| Subgrade | There are 35 subgrades for loans ranging from A1 down to G5, A1-subgrade is the safest while G5 is the riskiest. |
| Loan purpose | There are 14 purposes to borrow loan in Lending Club, namely, wedding, credit card, car loan, major purchase, home improvement, debt consolidation, house, vacation, medical, moving, renewable energy, educational, small business, and other. |
| FICO score | A measure of consumer credit score based on credit reports that range from 300 to 850. However, the FICO score in Lending Club ranges from 660 to 850 with step size of 5. |
| Annual income | The annual income of the borrower. |
| Housing situation | Own, rent, mortgage, and other are 4 levels that describe housing situation of the borrower |
| Employment length | The number of years the borrower has been working with his/her current employer (e.g. 1,2, 3, etc.) |
| Credit history length | The credit age of the borrower from the earliest credit trade line listed in credit report. |
| Delinquency 2 years | The number of delinquencies, i.e. more than 30 day past-due in the borrower's credit report for the past 2 years. |
| Inquiries last 6 months | The number of inquiries listed in borrower's credit report during the past 6 months. |
| Public records | Number of derogatory in borrower's credit report |
| Revolving utilization rate | The amount of credit the borrower is using out of all available revolving credit amount. |
| Open accounts | The number of open trade lines in the borrower's credit report |
| Months since last delinquency | The number of months from the borrower's last delinquency |
| Loan amount to annual income | Borrower's Loan amount over his/her annual income |
| Annual installment to income | The ratio of annual installment of the borrower over his/her annual income |
| Debt to income ratio (dti): | Monthly payments on the total debt of the borrower, excluding mortgage, divided by his/her monthly income. |

**4.1.2 Imbalanced class distribution**

Of the past loans in the Lending Club data, about 15% are default, and 85% are non-default[9]. This indicates the presence of class imbalance problem where non-default loans are the majority class and default loans are the minority class. A classifier trained on an imbalanced dataset often tends to ignore minority class while focusing on classifying the majority class accurately. Clearly this is very problematic in P2P lending, as misclassifying the default loans would be very expensive. Therefore, in order to accurately predict the default loans (as much as it is possible to do so), the class imbalance problem must be taken into account in PD prediction.

---

[9] Note that the imbalanced distribution of loans is common in other P2P lending platforms such as Prosper, and Kiva.

Class imbalance problem has been extensively studied in the literature (Japkowicz 2000; Japkowicz and Stephen 2002). Various techniques have been applied to address this issue such as re-sampling strategies, adjusting miss-classification costs, and adjusting the decision threshold $\gamma$, among which re-sampling strategies are the most popular ones. In this correspondence, we utilize re-sampling strategies to tackle the problem of class imbalance in training the PD predictive model. Re-sampling strategies aims to balance the class distribution using either undersampling or oversampling methods. Undersampling focuses on random elimination of majority class samples, while oversampling focuses on random replication of minority class samples. In addition to random replication, oversampling can also balance the class distribution by generating new samples on the minority class. Synthetic minority oversampling technique (SMOTE) forms new minority class samples by interpolating between several minority class samples that lie together (Chawla 2002). In this paper, random undersampling, random oversampling, and SMOTE are studied to evaluate the most effective re-sampling technique for balancing the class distribution of the loan status.

**4.1.3 Training the PD predictive model**

The procedures of training the predictive model is presented in Figure 7. The dataset is randomly split into training samples and test samples. Training samples include PD determinant features (e.g. borrower's assessment, borrower and loan characteristics) which are used to build the model. However, to address the problem of class imbalance, re-sampling techniques are used to balance the training samples. Finally, in order to achieve both memorization and generalization in PD prediction, wide and deep learning is used to build the model on the balanced training samples. Predictive performance of the trained model is then evaluated using the test samples. Once the trained model is finalized, it can be used as a classifier on new loans to identify non-default loans by PD prediction.

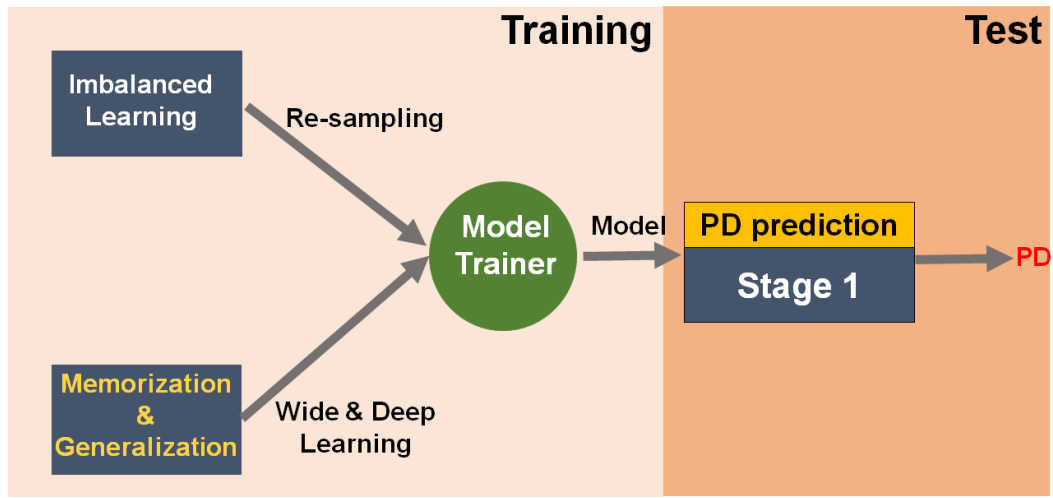

Figure 7. Training procedures for PD predictive model.

**4.2 Stage 2–IRR prediction**

**4.2.1 IRR determinant features**

The IRR features are exactly similar to PD features (see Table1). The IRR features are also similar to the features used in Serrano-Cinca, and Gutiérrez-Nieto (2016), therefore, it would be possible to conduct a fair comparison between the proposed model and the existing models. The above features form independent variables of the model, while IRR is the dependent variable in the model. The payment amount and the payment date of past loans are available in the payment data. Consequently, the IRR of the past loans can be easily computed using a common financial formula[10]. The distribution of IRR for the Lending Club loans is presented in Figure 1. From this figure, the IRR distribution is skewed to positive values (almost 85% of the past loans have positive IRR). This is consistent with the loan status distribution showing 85% non-default loans. Indeed the IRR skewness is originally caused by the fact that whether borrowers successfully pay/default the loans. Hence, the IRR prediction also suffers from the class imbalance problem.[11] In our proposed two-stage approach, stage 1 already tackles this imbalance problem by using re-sampling techniques. Hence, in stage 2, it would not be an issue anymore, as the IRR predictive model is only built on the records with positive IRR. As a result, the data used for training the IRR prediction model is not skewed (please refer to Figure 1).

---

[10]Lending Club's payments data are available at http://additionalstatistics.lendingclub.com. The payment data should be joint with the loan data for this analysis.
[11]

**4.2.2 Training the IRR predictive model**

The procedures of training the IRR predictive model is presented in Figure 8. The training samples used for stage 2 is exactly the same as stage 1, but with an exception that only samples with positive IRR are taken into account. The reason is that by evaluating the loan's PD predicted in stage 1, the status of the loan can be identified. In case the loan is predicted as non-default, then it would be expected that the loan would result into positive IRR. Hence, there is no need to train the IRR model on the training samples with negative IRR, as it degrades the prediction accuracy (more details are discussed in Section 5 where the proposed model is compared with profit scoring approach of Serrano-Cinca and Gutiérrez-Nieto (2016) that is trained on the training samples with both positive and negative IRR).

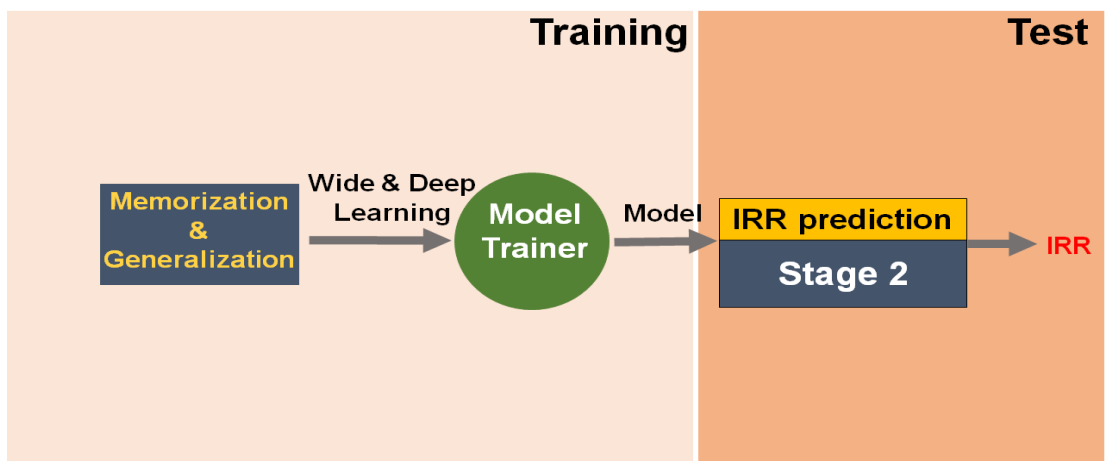

Figure 8. Training procedures for the IRR predictive model.

In order to achieve both memorization and generalization in the IRR prediction, wide and deep learning is used to build the model on training samples with positive IRR. Predictive performance of the trained model is then evaluated using the same test samples as stage 1 (again only samples with positive IRR are tested). The trained IRR model in stage 2 is then used along with the trained PD model in stage 1 to score the loans. As specified, the trained PD model identifies the loans status, and for the loans that are predicted as non-default, the IRR is subsequently predicted using the trained IRR model. Finally, the loans with highest IRR can be selected for lender's investment.

**5. Numerical studies**

The numerical studies in this paper are based on the Lending Club data (similar to other studies such as Serrano-Cinca and Gutiérrez-Nieto 2016). All information about the borrowers and their payments are available online. The dataset includes borrower's information from 2007. Similar to Serrano-Cinca, and Gutiérrez-Nieto (2016), loans in 2007 were removed from the analysis as they were issued under the company's pilot model. The loan terms are typically 3 or 5 years. Therefore, the loans issued up to the end of 2013 were included in our analysis as the IRR can be calculated for the loans whose term length is finished. PD features and IRR features are captured from the data, and are split into a training set and a test set with proportion of 80% and 20%, respectively. The training set is used to build the proposed PD and IRR predictive models, and the test set is used to evaluate the performance of the proposed models.

**5.1 PD predictive model – the Lending Club case**

**5.1.1 Implementation of resampling techniques to balance the training data**

The first numerical studies pertain to the PD predictive modeling. As discussed in Section 4.1 and presented in Figure 7, there are two computational tasks involved, namely, re-sampling, and wide and deep learning. The re-sampling techniques were implemented using "imbalanced-learn" API in Python.[12] In this Python API, numerous re-sampling techniques are available, among which we implement the most popular techniques for comparison purpose, namely, (1) random undersampling, (2) random oversampling, and (3) SMOTE. Random undersampling balances the distribution of default loan and non-default loan classes by random elimination of non-default loan samples. Using random undersampling the number of training samples on each class will be around 12,000. Random oversampling balances default loan and non-default loan distributions by random replication (with replacement) of default loan samples. Using random oversampling, the number of training samples on each class will be 64,000. SMOTE balances the classes by generating new synthetic samples on default loans. Synthetic samples are created in the following way. The distance of each default loan sample and its k-nearest neighbor samples are calculated. This distance is multiplied by a random number (between 0 and 1) and is added to the under consideration sample. The result is called a synthetic sample. There is about 50,000 oversampling needed. Hence from all synthetic samples created by all the k-nearest neighbors, about 50,000 are selected. Using SMOTE, the number of training samples on each class will be 64,000. Note that the SMOTE algorithm has a tuning parameter k, which should be specified through the numerical studies. The default value is 5.

---

[12] http://contrib.scikit-learn.org/imbalanced-learn/index.html

**5.1.2 Implementation of the wide and deep learning for the PD prediction**

The second calculation task is to implement the wide and deep learning model on the training samples which were balanced by re-sampling in Section 5.1.2. Tensorflow API in Python is utilized for implementing wide and deep learning.[13] The implementation requires identifying the wide components and the deep components. The wide components include the categorical features including basis features and engineered features. The wide components are presented in Table 2. The engineered features include the interaction terms of some pairs of basis features, in order to capture memorization. For example, interaction term AND (FICO, purpose) of the FICO score and the loan purpose, and interaction term AND (housing, Emp.) of the housing situation and the employment length are defined using Tensorflow API.

Table 2. Wide components used in the wide and deep learning model. The wide components include categorical basis features and engineered features including interaction terms.

| Wide components | Raw/Basis Features | Engineered Features |
| --- | --- | --- |
| Grade | ✓ | |
| Subgrade | ✓ | |
| Loan purpose | ✓ | |
| FICO score | ✓ | |
| Housing situation | ✓ | |
| Employment length | ✓ | |
| AND(FICO, purpose) | | ✓ |
| AND (subgrade, FICO) | | ✓ |

---

[13] Tensorflow is an open-source software library package for machine learning (especially deep learning) which was originally developed by the scientists on the Google Brain team, and was released in November 2015. It is currently being used for research and production in Google products.

The deep components include continuous features including both real-valued basis features and embeddings of the categorical basis features. Table 3 shows the deep components. The real-valued embeddings are the transformation of the categorical features from their original space into low dimensional real-valued space. The embedding vectors enable generalization in the PD prediction. The real-valued embeddings of the sub-grade, the purpose and the FICO score are defined using Tensorflow API (the dimension of the embedding vectors is set to 8 as we obtained best performance with this dimension). It should be noted that FICO is an ordinal variable therefore, mapping FICO into embedding vectors will lose the ordinal information. However, it would not be a problem in lending club data analysis as the hypothesis that higher FICO score always results into lower default rates is not valid. Please refer to Appendix II for more details.

Table 3. Deep components used in the wide and deep learning model. The deep components include continuous basis features and engineered features including embeddings from categorical basis features.

| Deep components | Raw/Basis Features | Engineered Features |
|---|---|---|
| Annual income | ✓ | |
| Credit history length | ✓ | |
| Delinquency 2 years | ✓ | |
| Inquiries last 6 months | ✓ | |
| Public records | ✓ | |
| Revolving utilization rate | ✓ | |
| Open accounts | ✓ | |
| Months since last delinquency | ✓ | |
| Loan amount to Inc. | ✓ | |
| Annual installment to Inc. | ✓ | |
| Debt to Inc. ratio (dti): | ✓ | |
| Subgrade embedding | | ✓ |
| Purpose embedding | | ✓ |
| FICO embedding | | ✓ |

After introducing the wide and deep components, the structure of the model should be defined. The model (deep model) consists of three hidden layers with 100, 50, and 10 activations (neurons) in first, second and third layer, respectively[14]. ReLU was used as the activation function in all the hidden layers. Gradient descent algorithm with learning rate 0.002 was used to optimize the loss function defined as cross-entropy. All the above procedures can be easily programed in a few lines of code utilizing the Tensorflow API.

Training the wide and deep learning model for the PD prediction was conducted in 1,000 steps with batch size of 100. The training is carried out under the supervised settings where the actual outputs are known (i.e. either default loan or non-default loan). In each step 100 samples (without replacement) are randomly selected from the training data and the model parameters (i.e. weights and biases for both deep and wide components as well as the embedding vectors) are learned in a way to minimize the average loss function that evaluates the difference of the actual output and the model output for each instance in the batch. To avoid overfitting dropout approach with rate of 0.2 is used in the training (Sirvastava et al. (2012)). The optimization procedures described above are adopted from stochastic gradient descent algorithm (the readers are referred to Bottou (2010) for more details). Figure 9 shows how the model is trained over 1000 steps. It can be observed that the loss function is reduced in each step. This shows how effective the optimization procedures are in training the wide and deep learning model. The optimization process was stopped after 1000 iterations as the performance does not improve further. The validation set is also used to verify the stopping criteria for optimization procedures. The highest PD prediction accuracy achieved at 1000 steps and does not improve significantly after 1000 steps.

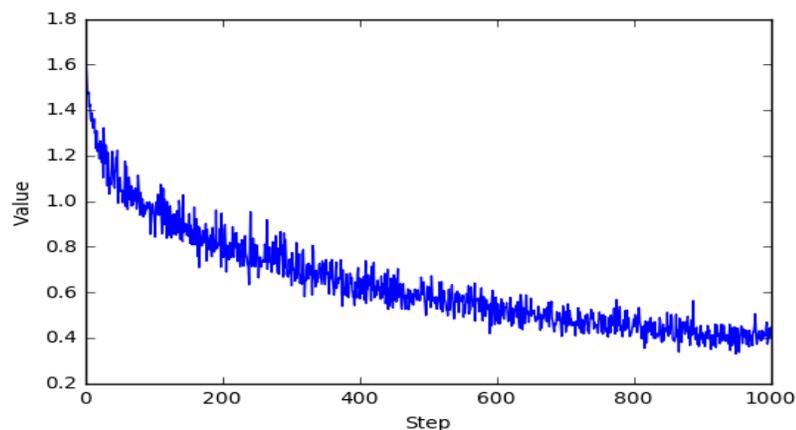

Figure 9. Learning the PD predictive model by minimizing cross-entropy loss function in 1,000 steps.

---

[14] The structure of the model can be verified using numerous experiments to evaluate which combination could provide higher predictive performance for the model.

**5.1.3 Evaluating the performance of PD model on the Lending Club test samples**

After training the model, it is crucial to evaluate the performance of the model on the test samples and compare its performance with the benchmarks, namely, wide learning (logistic regression), and deep learning (neural network)[15]. Precision and recall are used as performance metrics of the models. It is well-known that in the presence of imbalanced dataset, accuracy is not a proper performance metric, and precision and recall are typically preferred. The precision and recall for each of the classes are defined in Table 4.

Table 4. Performance metrics utilized to evaluate the performance of the PD prediction

| True | Prediction | |
|---|---|---|
| | Non-Default | Default |
| Non-Default | True Positive (TP) | False Negative (FN) |
| Default | False Positive (FP) | True Negative (TN) |

$$\text{Precision\_Positive} = \frac{TP}{TP+FP}, \quad \text{Recall\_Positive} = \frac{TP}{TP+FN}$$
$$\text{Precision\_Negative} = \frac{TN}{TN+FN}, \quad \text{Recall\_Negative} = \frac{TN}{TN+FP}$$

Using the performance metrics, Table 5 compares the performance of all these three models in combination with the three re-sampling techniques used to balance the training samples. The results indicate that the combination of SMOTE and wide and deep learning leads to the highest Precision_P and Recall_N. In general it can be observed that SMOTE leads to the highest performance regardless of which algorithm is used for the predictive modeling step.

---

[15] Both wide learning and deep learning models were implemented in Tensorflow. Note that each of these models are trained on the same training set. The best trained model on each of these algorithms are then used for comparison.

Table 5. Performance comparison of the wide and deep learning model versus the wide learning, and the deep learning models in identifying loan status using the PD prediction.

| Resampling | Wide Learning | |
|---|---|---|
| | Precision_P | Recall_N |
| Under Sampling | 0.75 | 0.62 |
| Over Sampling | 0.78 | 0.65 |
| SMOTE* | 0.86 | 0.64 |
| | **Deep Learning** | |
| Under Sampling | 0.77 | 0.61 |
| Over Sampling | 0.81 | 0.64 |
| SMOTE* | 0.88 | 0.69 |
| | **Wide & Deep Learning** | |
| Under Sampling | 0.85 | 0.67 |
| Over Sampling | 0.89 | 0.65 |
| SMOTE* | 0.91 | 0.74 |

**5.2 IRR predictive model – the Lending Club case**

**5.2.1 Implementation of the wide and deep learning for the IRR prediction**

The second numerical study relates to the IRR prediction. From Figure 8, there is only one computational task remained, namely, training the wide and deep learning model for the IRR prediction. As discussed in Section 4.2.2, the training samples with positive IRR will be considered for training the model. Hence, no class imbalance problem exists in the training data in this stage.

Similar to Section 5.1.2, the wide and deep components are defined in Table 2 and Table 3, respectively. The structure of the deep model is similar to that of the deep model in the PD prediction in Section 5.1.2. This model consists of three hidden layers with 100, 50, and 10 activations (neurons) in first, second and third layer, respectively. ReLU was used as the activation function in all the hidden layers. Gradient descent algorithm with learning rate 0.002 was used to optimize the loss function defined as mean squared error. All the above procedures can be easily programed in a few lines of code in the Tensorflow API.

Training the wide and deep learning model for the IRR prediction was conducted in 1,000 steps with batch size 100. Figure 10 shows how the IRR model is trained over 1,000 steps. This shows how effective the optimization procedures are in training the wide and deep learning model for the IRR prediction.

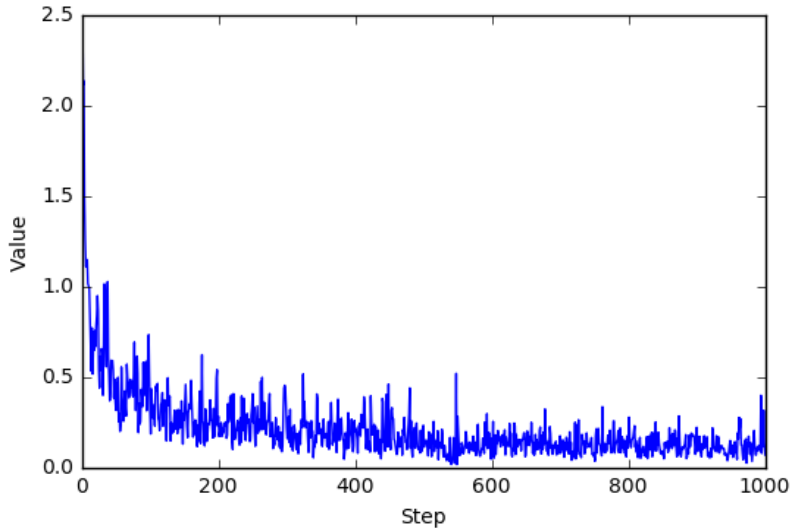

Figure 10. Learning the IRR predictive model by minimizing mean squared error loss function in 1,000 steps.

**5.2.2 Evaluating the performance of the IRR model on the Lending Club test samples**

We further conduct evaluation on the performance of the trained IRR model on the test samples (again samples with positive IRR are considered for the test) and compare its performance with the benchmarks, namely, wide learning (multivariate linear regression), and deep learning (neural network). Mean squared error (MSE) is used as a performance metric, which is very common in regression problems. Table 6 compares the performance of all these three models. Wide and deep learning obtains the lowest MSE which indicates the effectiveness of the proposed approach for the IRR prediction.

Table 6. Performance comparison of the wide and deep learning model versus the wide learning, and the deep learning models in the IRR prediction

| Model | MSE |
|---|---|
| Wide | 0.078 |
| Deep | 0.071 |
| Wide & Deep | 0.068 |

**5.3. Two-stage scoring approach versus credit and profit scoring approaches**

In Sections 5.1 and 5.2, the implementation procedures of building the PD and the IRR models were presented. The proposed two-stage scoring approach utilizes these two models to identify the best loans for the investors. Therefore, it would be crucial to evaluate the overall performance of the proposed two-stage approach, and compare its performance with the existing credit and profit scoring approaches. As specific, three approaches are compared, namely a credit scoring approach based on logistic regression (Approach 1), the profit scoring appraoch proposed in Serrano-Cinca and Gutiérrez-Nieto (2016) based on decision tree (Approach 2), and our proposed two-stage scoring approach (Approach 3). The output of Approach 1 is the PD of the test loans, and the output of Approach 2 and Approach 3 is IRR of the test loans[16]. Given the approaches' outputs (scores), the lenders select the loans with the highest scores. For the comparison purpose, we consider a scenario that a lender chooses 30 best loans according to the scores of these three approaches. Note that the output of Approach 1 is the PD. Hence, a lender selects the loans with the lowest predicted PD. On the other hand, the outcome of Approach 2 and Approach 3 is the IRR. Hence, a lender chooses the loans with the highest predicted IRR. The profitability of a lender given these three approaches are calculated in terms of average IRR (over the 30 best loans) and reported in Table 7.

The results indicate profitability of our proposed approach (Approach 3) is superior over Approach 2 and Approach 1. It was expected that Approach 1 would result into the lowest profitability as it is focused on PD rather than profitability. The performance of Approach 3 is much better than Approach 2 mainly because of the fact that our proposed approach is capable of addressing the class imbalance problem that was completely ignored in Approach 2. In order to emphasize more on the contribution of the proposed approach, Figure 11 is presented to visually compare predictive power of Approach 3 over that of Approach 2. In this figure, both vertical and horizontal axes are the actual IRR of the test data with positive values. The blue squares show the actual IRR that are located in the plots with a slope of 45 degree). The predicted IRR are plotted with orange squares. Figure 11 belongs to Approach 3 (left plot) and Approach 2 (right plot), respectively.

---

[16] Here, our proposed approach is applied on the whole set of test samples. Previously in Section 5.2.3 to test the IRR prediction power in stage 2 (Section 5.2.3), we only used the test samples with positive IRR. However, in this comparison all the test samples should be used to evaluate and compare the overall performance of our two-stage scoring approach.

Table 7. Average IRR resulted from using Approach 1, Approach 2, and Approach 3.

| Approach 3 | Approach 2 | Approach 1 |
|---|---|---|
| 0.1619 | 0.0780 | 0.0631 |

The ideal case happens if the predicted IRR and actual IRR are exactly equal, which means that the orange squares are exactly located on the blue squares and form almost a line with a 45 degree slope. From our visualization, it can be understood how accurate Approach 3 is in the IRR prediction because the orange squares are located very close to the blue squares. However, this is not the case in Approach 2 as the orange squares are highly spread around the blue squares. Figure 11 justifies why Approach 3 obtained much higher profitability than Approach 2. Addressing the loan status imbalance problem, Approach 3 is able to accurately predict the test samples with positive IRR which is not possible when using Approach 2. This means that the lenders would be able to choose the loans whose actual IRR are high (e.g. larger than 20%) and the proposed approach is able to correctly predict high IRR for those loans.

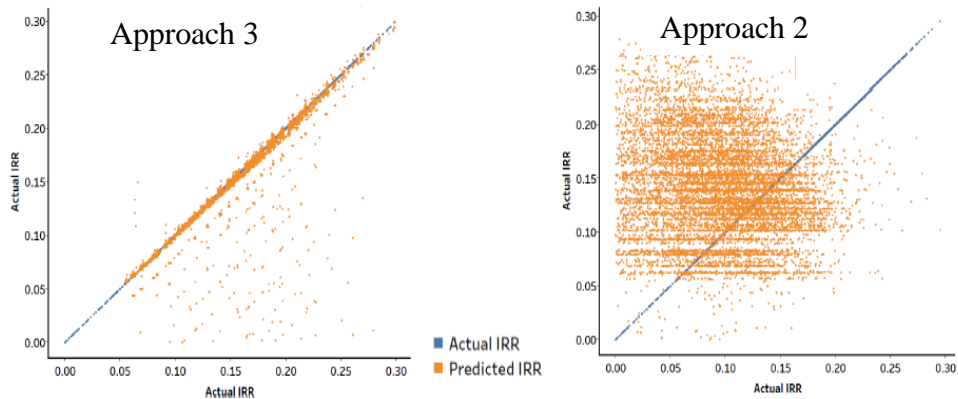

Figure 11. Visualization of actual positive IRR versus predicted IRR by Approach 3, and Approach 2.

Scoring approaches such as Approach 1, Approach 2 and Approach 3 allow lenders to decide their fund allocations in the P2P lending market. Depending on the lenders' investment goals, any of the above approaches could become handy. To elaborate more on this point, Table 8 presents detailed information on a few of best loans (based on their scores) identified by these three approaches. Along with the score, the actual IRR and the grade of each of these loans are illustrated. All the best loans identified by Approach 1 have very low actual IRR (in the range of 5%-7%) and were assigned Grade A by the Lending Club. Although, the profitability performance of Approach 1 would be relatively low, it rarely resulted into the loans that were defaulted. Hence, for conservative lenders whose investment goal is to avoid riskier loans as much as possible, and achieve a fair amount of profit, Approach 1 could help them in their fund allocation decisions.

Table 8. Actual IRR and the Lending Club grade reported for a few of best loans identified by Approach 1, Approach 2, and Approach 3

| Approach 3 | | | Approach 2 | | | Approach 1 | | |
|---|---|---|---|---|---|---|---|---|
| Grade | Actual | Score | Grade | Actual | Score | Grade | Actual | Score |
| F | 0.2745 | 0.2764 | B | -0.2644 | 0.2095 | A | 0.0624 | 0.9586 |
| F | 0.2693 | 0.2717 | G | 0.2593 | 0.1779 | A | 0.0626 | 0.958 |
| F | 0.2625 | 0.2657 | C | 0.1679 | 0.1708 | A | 0.0621 | 0.9576 |
| F | 0.2813 | 0.2648 | E | 0.2387 | 0.1704 | A | 0.0636 | 0.9558 |
| E | 0.223 | 0.2606 | B | 0.1283 | 0.1679 | A | 0.0624 | 0.9534 |
| F | 0.2463 | 0.2598 | C | 0.1633 | 0.1626 | A | 0.0628 | 0.953 |
| E | -0.1924 | 0.2574 | B | -0.7704 | 0.1604 | A | 0.0622 | 0.9526 |
| E | -0.4682 | 0.2529 | B | 0.1326 | 0.1599 | A | 0.0571 | 0.95 |
| E | 0.2522 | 0.2529 | C | 0.1459 | 0.1599 | A | 0.0633 | 0.9494 |
| E | -0.602 | 0.2474 | B | -0.0597 | 0.1589 | A | 0.0691 | 0.9492 |
| E | 0.247 | 0.2468 | D | 0.203 | 0.1584 | A | 0.0626 | 0.9484 |
| E | -0.1623 | 0.2433 | D | 0.176 | 0.1542 | A | 0.0559 | 0.9472 |
| E | 0.2479 | 0.2414 | C | 0.1538 | 0.154 | A | 0.0694 | 0.945 |
| E | 0.2387 | 0.2408 | C | 0.1534 | 0.1538 | A | 0.0625 | 0.943 |
| E | 0.2403 | 0.2394 | C | 0.1678 | 0.1527 | A | 0.0616 | 0.9414 |
| E | 0.2374 | 0.239 | B | 0.1404 | 0.1526 | A | 0.0626 | 0.9402 |
| E | 0.2468 | 0.2379 | C | 0.1468 | 0.1493 | A | 0.0682 | 0.9402 |
| E | 0.2334 | 0.2358 | D | 0.208 | 0.1474 | A | 0.0702 | 0.9398 |

The best loans identified by Approach 2 have actual positive IRR in a wide range of (12%,25%). Although, the profitability of Approach 2 is higher than Approach 1, it could result into defaulted loans. More importantly, loans identified by Approach 2 compose of various grades (including B,C,D,E and G). This indicates that a portfolio of loans with various risk levels can be selected by this model. Therefore, for the lenders with balanced investment strategy Approach 2 could be a better option. The best loans identified by Approach 3 include high actual positive IRR in a range of (22%,28%). Consequently, the profitability of Approach 3 is the highest among all models presented here. Similar to Approach 2, Approach 3 may also lead to default loans. More importantly, the loans selected by Approach 3 compose of high risk grades including E and F. Hence, for the aggressive lenders whose investment goal is to select highly profitable portfolios that typically include riskier loans (e.g. Grade E, F, and G), Approach 3 would be the best option.

**6. Conclusion**

This paper proposed a two-stage scoring approach to help lenders choose the best loans for investment in the P2P lending market. The proposed approach is developed to address the challenges with the existing scoring approaches, i.e., credit scoring and profit scoring. Credit scoring focuses on PD prediction and identifies the best loans as the ones with the lowest PD. Unfortunately, credit scoring completely ignores to deliver the main need of lenders which is prediction of profit they may receive through their investment. On the other hand, profit scoring is able to satisfy that need by investment profitability prediction. However, profit scoring totally ignores the class imbalance problem. Consequently, ignorance of the class imbalance problem significantly affects the accuracy of profitability prediction. Our proposed two-stage scoring approach is an integration of credit scoring and profit scoring to tackle the above challenges. More specifically, the first stage was designed as credit scoring to identify non-default loans while imbalanced nature of loan status was taken into account in the PD prediction. The loans identified as non-default were then moved to stage 2 for profitability prediction (IRR). Wide and deep learning was used to build the predictive models in both stages of the proposed approach, to achieve both memorization and generalization.

The proposed approach was verified using real-world data from Lending Club which is one of the largest P2P lending platforms in the US. In Stage 1, the class imbalance problem of the training dataset was tackled using re-sampling techniques such as random undersampling, random oversampling, and SMOTE. Tensorflow was then utilized for wide and deep learning implementation to build the PD predictive model using the balanced training samples. The performance of the model was evaluated based on test samples. The test results indicated that the wide and deep learning model in combination with SMOTE achieves the highest performance on the loan status prediction in comparison with the benchmark algorithms including wide learning and deep learning. In Stage 2, the IRR predictive model was built using wide and deep learning (in Tensorflow) based on the training samples with positive IRR values. Test samples with positive IRR were also utilized to evaluate the performance of the model; the test results indicated that the wide and deep learning IRR predictive model was superior over other benchmarks. The trained predictive models in Stage 1 and Stage 2 were integrated to form the proposed two-stage scoring approach. Finally, overall performance of the two-stage scoring approach was evaluated using the test samples (all test samples were included i.e. with positive IRR or negative IRR). It was shown that the two-stage scoring approach achieves the highest profitability compared to the credit scoring approach using logistic regression and the profit scoring approach using decision tree. As specific, if the lenders select the top 30 best loans identified by each of these approaches, they would be able to obtain average IRR of 16%, 7%, and 6% using two-stage scoring, profit scoring, and credit scoring approaches, respectively.

The variables used in this study are all collected from Lending Club public dataset. However, there must be other personal variables on the borrowers that are not listed in the public data. These variables include personal level information such as demographics, social media information, home addresses, job titles (Hunt et al. 2018). These features would certainly provide much granular information on the consumer and consequently improve the performance of profit scoring. For example zip code and block group of the consumers can be used as a feature in the model; each are categorical variables with more than 30000 levels. Embedding vectors through deep learning on these categorical variables would improve the generalization ability of the model. The interaction terms through wide learning would bring memorization into the learning procedures. Therefore, this makes the importance of the proposed two-stage scoring approach readily apparent. As the proposed approach uses wide and deep learning model then it can effectively integrate both memorization and generalization in the learning procedures. In the future, the utility of the proposed scoring approach in the presence of more granular variables can be explored.

**References**

Asgari, E. and Bastani, K., 2017. The Utility of Hierarchical Dirichlet Process for Relationship Detection of Latent Constructs. In Academy of Management Proceedings (Vol. 2017, No. 1, p. 16300). Briarcliff Manor, NY 10510: Academy of Management.
Banasik, J., Crook, J. N., & Thomas, L. C. (1999). Not if but when borrowers default. Journal of Operational Research Society 50, 1185–1190.
Bottou, L., 2010. Large-scale machine learning with stochastic gradient descent. In Proceedings of COMPSTAT'2010 (pp. 177-186). Physica-Verlag HD.
Buckley, J., & James, I. (1979). Linear regression with censored data. Biometrika 66, 429–436.
Breslow, N.E., 1975. Analysis of survival data under the proportional hazards model. *International Statistical Review/Revue Internationale de Statistique*, pp.45-57.
Bradburn, M.J., Clark, T.G., Love, S.B. and Altman, D.G., 2003. Survival analysis part II: multivariate data analysis–an introduction to concepts and methods. *British journal of cancer*, *89*(3), p.431.

Chawla, N.V., Bowyer, K.W., Hall, L.O. and Kegelmeyer, W.P., 2002. SMOTE: synthetic minority over-sampling technique. Journal of artificial intelligence research, 16, pp.321-357.

Cheng, H.T., Koc, L., Harmsen, J., Shaked, T., Chandra, T., Aradhye, H., Anderson, G., Corrado, G., Chai, W., Ispir, M. and Anil, R., 2016, September. Wide & deep learning for recommender systems. In Proceedings of the 1st Workshop on Deep Learning for Recommender Systems (pp. 7-10). ACM.

Emekter, R., Tu, Y., Jirasakuldech, B. and Lu, M., 2015. Evaluating credit risk and loan performance in online Peer-to-Peer (P2P) lending. Applied Economics, 47(1), pp.54-70.

Finlay, S., 2010. Credit scoring for profitability objectives. European Journal of Operational Research, 202(2), pp.528-537.

Finlay, S.M., 2008. Towards profitability: A utility approach to the credit scoring problem. Journal of the Operational Research Society, 59(7), pp.921-931.

Guo, Y., Zhou, W., Luo, C., Liu, C. and Xiong, H., 2016. Instance-based credit risk assessment for investment decisions in P2P lending. European Journal of Operational Research, 249(2), pp.417-426.

Hunt, R.A., Townsend, D.M., Asgari, E. and Lerner, D.A., 2018. Bringing It All Back Home: Corporate Venturing and Renewal through Spin-ins. Entrepreneurship Theory and Practice, p.1042258718778547.

Japkowicz, N., 2000, June. The class imbalance problem: Significance and strategies. In Proc. of the Int'l Conf. on Artificial Intelligence.

Japkowicz, N. and Stephen, S., 2002. The class imbalance problem: A systematic study. Intelligent data analysis, 6(5), pp.429-449.

Jiang, C., Wang, Z., Wang, R. and Ding, Y., 2018. Loan default prediction by combining soft information extracted from descriptive text in online peer-to-peer lending. Annals of Operations Research, 266(1-2), pp.511-529.

Lai, T. L., & Ying, Z. L. (1994). A Missing information principle and M-estimators in regression analysis with censored and truncated data. Annals of Statistics 22, 1222–1255.

Lessmann, S., Baesens, B., Seow, H.V. and Thomas, L.C., 2015. Benchmarking state-of-the-art classification algorithms for credit scoring: An update of research. European Journal of Operational Research, 247(1), pp.124-136.

Malekipirbazari, M. and Aksakalli, V., 2015. Risk assessment in social lending via random forests. Expert Systems with Applications, 42(10), pp.4621-4631.

Narain, B. (1992). Survival analysis and the credit granting decision. In: Thomas, L. C., Crook, J. N., & Edelman, Decision Sciences 28, 105–120. D. B. (Eds.), Credit scoring and credit control, Oxford Thomas, L. C. (1992). Financial risk management models. University Press, Oxford, pp. 109–122.

Sanchez-Barrios, L.J., Andreeva, G. and Ansell, J., 2016. Time-to-profit scorecards for revolving credit. *European Journal of Operational Research*, *249*(2), pp.397-406.

Serrano-Cinca, C., Gutiérrez-Nieto, B. and López-Palacios, L., 2015. Determinants of default in P2P lending. PloS one, 10(10), p.e0139427.

Serrano-Cinca, C. and Gutiérrez-Nieto, B., 2016. The use of profit scoring as an alternative to credit scoring systems in peer-to-peer (P2P) lending. Decision Support Systems, 89, pp.113-122.

Srivastava, N., Hinton, G., Krizhevsky, A., Sutskever, I. and Salakhutdinov, R., 2014. Dropout: a simple way to prevent neural networks from overfitting. The Journal of Machine Learning Research, 15(1), pp.1929-1958.

So, M. C., Thomas, L. C., Seow, H. V., & Mues, C. (2014). Using a transactor/revolver scorecard to make credit and pricing decisions. Decision Support Systems, 59, 143-151.

Stewart, R.T., 2011. A profit-based scoring system in consumer credit: making acquisition decisions for credit cards. Journal of the Operational Research Society, 62(9), pp.1719-1725.

Thomas, L.C., Allen, D.E. and Morkel-Kingsbury, N., 2002a. A hidden Markov chain model for the term structure of bond credit risk spreads. International Review of Financial Analysis, 11(3), pp.311-329. Thomas, L.C., Edelman, D.B. and Crook, J.N., 2002b. Credit scoring and its applications. Society for Industrial and Applied Mathematics: Philadelfia. Monographs on mathematical modeling and computation, pp.89-106.

Verbraken, T., Bravo, C., Weber, R. and Baesens, B., 2014. Development and application of consumer credit scoring models using profit-based classification measures. European Journal of Operational Research, 238(2), pp.505-513.

**Appendix I.**

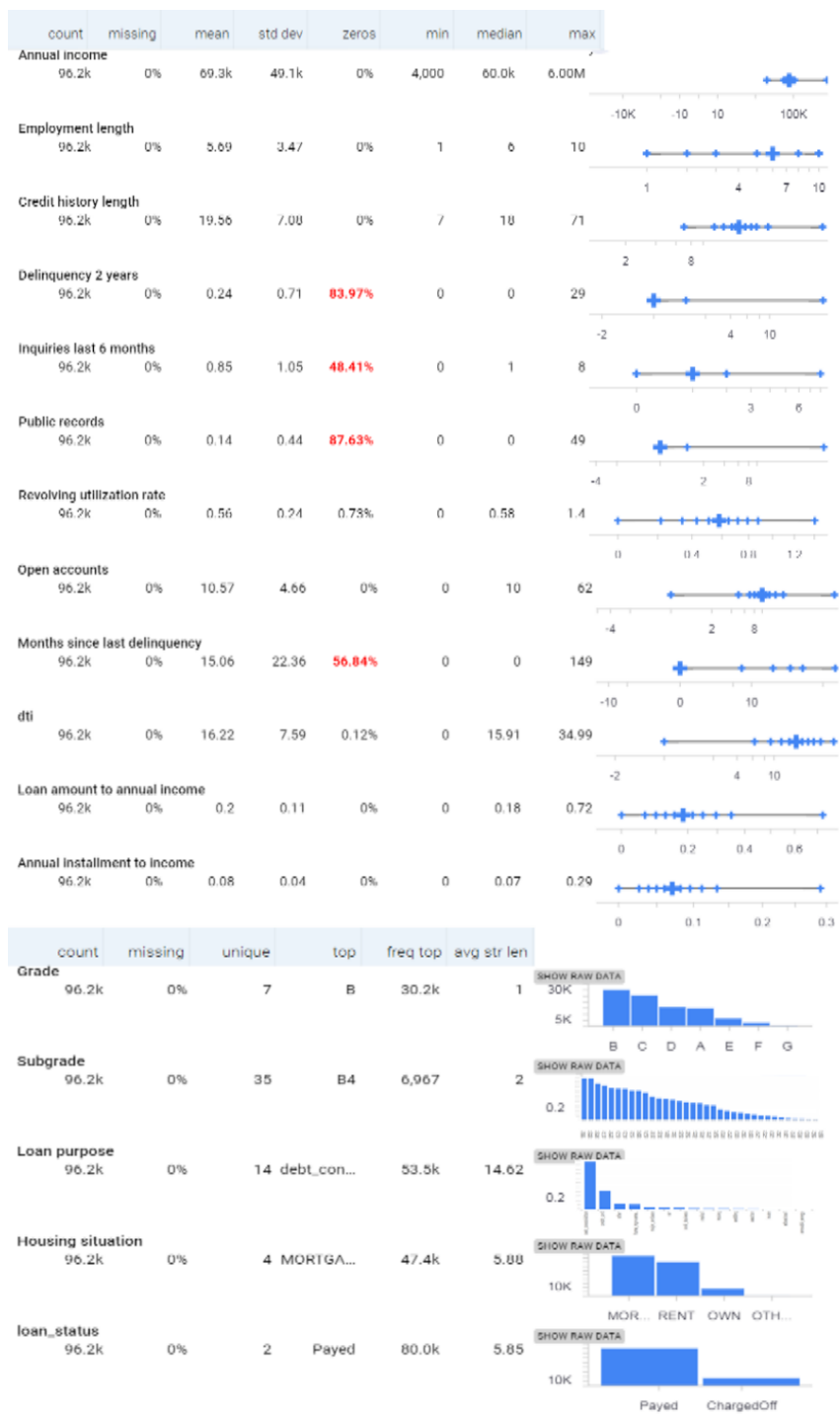

Figure A1. Descriptive statistics of the features used in PD and IRR predictive models.

**Appendix II.**

FICO is considered as a categorical variable in this work. Although it is an ordinal variable and mapping it into embedding vectors might result into loss of ordinal information, our analysis indicates that the ordinal nature of FICO score is not always important to be preserved. There are some scenarios that higher FICO score does not necessarily reduce the risk of default or increase profitability. Therefore, it is not important that the ordinal information is lost during learning FICO embedding vectors. To elaborate this we utilize Figure A2. This figure is generated using Facets (https://pair-code.github.io/facets/) which is an open source project of Pair. Pair (https://ai.google/research/teams/brain/pair) is part of Google Brain research team which is focused on enhancing research and design of people-centric AI systems. Using Facets we would be able to automatically visualize features, and understand machine learning datasets.

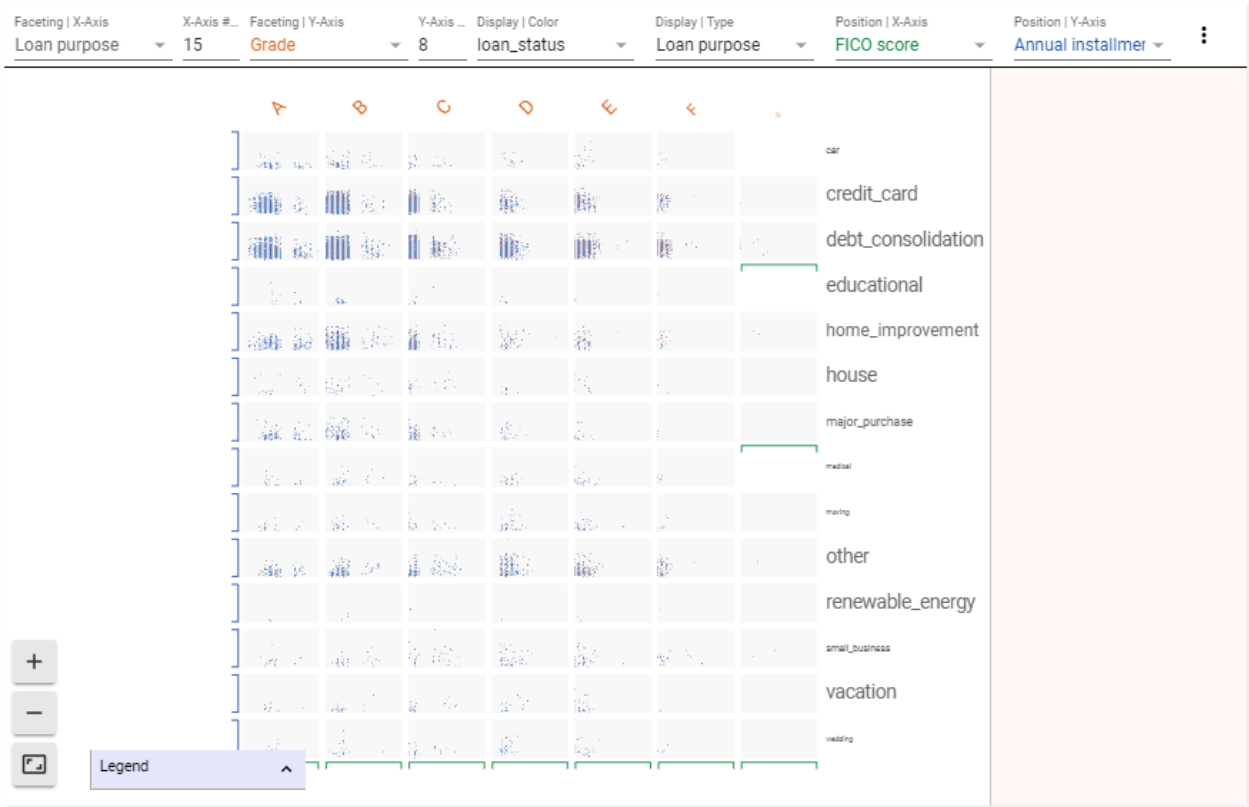

Figure A2. Interaction of four features namely grade, loan purposes, FICO score and annual installment to income. Loan status is used to color the data points (blue color shows payed loans and red color shows defaulted loans).

In this figure four features, namely, loan purpose, grade, FICO score, and annual installment to income are visualized. Loan status is used to color the data points (blue color shows payed loans and red color shows defaulted loans). We focus on data points with loan purposes of debt consolidation and credit card as they cover almost 77% of the distribution. The part of figure related to these two loan purposes is separately presented in Figure A3. In figure below green axis (horizontal axis) is the FICO score and blue axis (vertical axis) represents the annual installment to income. From Figure A3 it can be seen higher FICO score does not necessarily reduce the risk of default. For instance on Grade G and loan purpose of "debt consolidation", higher FICO score increases the risk of default (please refer to the black dashed circle at the bottom right of Figure A3).

Indeed, the important note here is that there is a complex non-linear relationship between dependent variables, i.e., IRR and PD with respect to the features. Wide and deep model learns the hidden layer weights as well as embedding weights of categorical variables such as FICO in a way to represent all these complex nonlinear relationships. Therefore, mapping an ordinal variable such as FICO into embedding vectors would not be a problem in our analysis. In fact, it allows generalization to the scenarios that are missing in the training but might occur in the production phase.

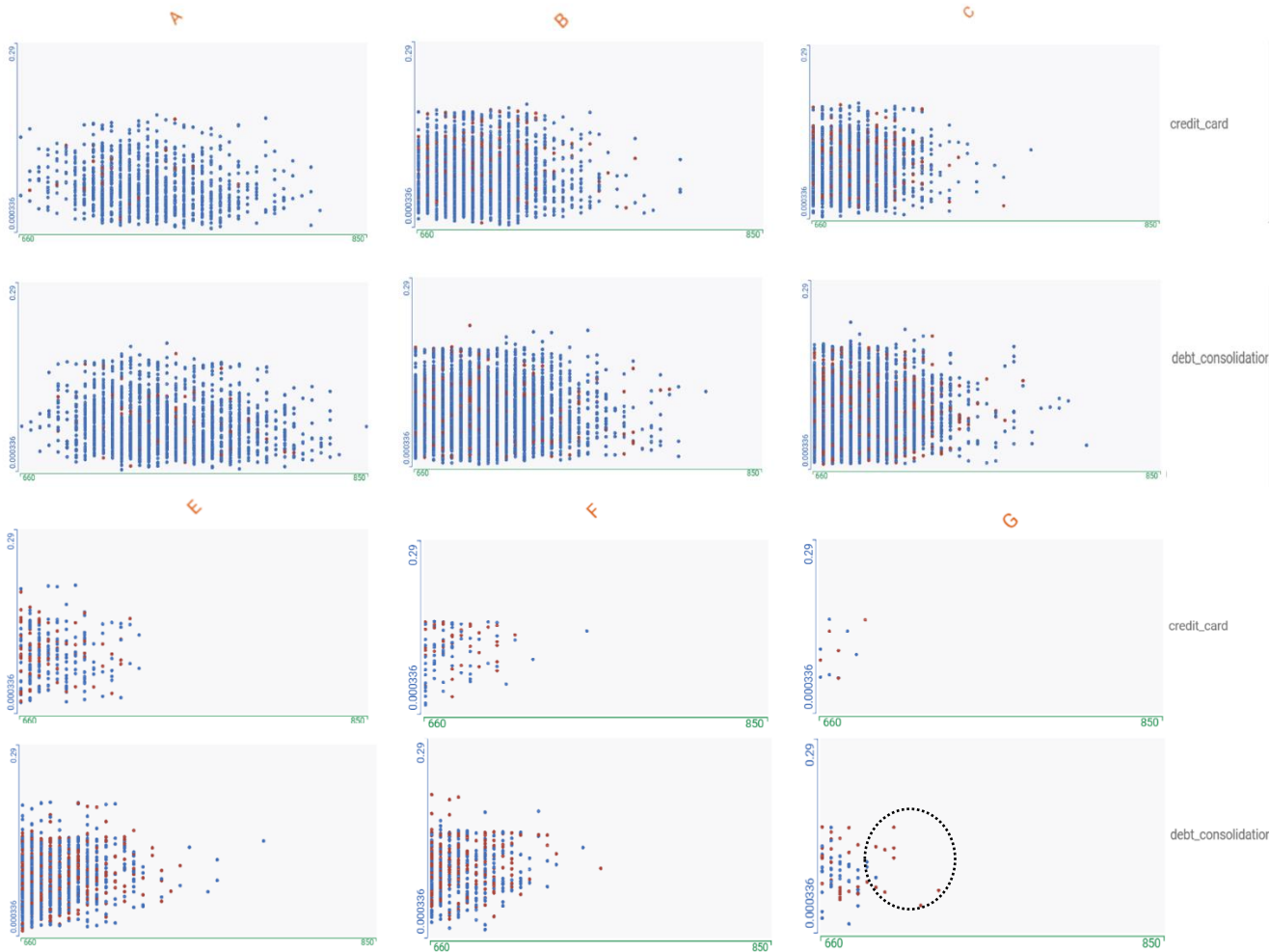

Figure A3. Interaction of "credit card" and "debt consolidation" loan purpose with grade, FICO score, and annual installment to income.